\pdfoutput=1
\documentclass[11pt]{article}
\usepackage[final]{acl}
\usepackage{latexsym}
\usepackage[T1]{fontenc}
\usepackage[utf8]{inputenc}
\usepackage{graphicx}
\usepackage{subfig}

\title{Conversational Agents and the Understanding of Human Language: Reflections on AI, LLMs, and Cognitive Science}

\author{
 Andrei Popescu-Belis\textsuperscript{1,2}\\
 \textsuperscript{1}\,HEIG-VD~/~HES-SO, 1401 Yverdon-les-Bains, Switzerland\\
 \textsuperscript{2}\,EPFL, 1015 Lausanne, Switzerland\\
   \texttt{andrei.popescu-belis@heig-vd.ch}
}

\begin{document}
\maketitle
\begin{abstract}
In this paper, we discuss the relationship between natural language processing by computers (NLP) and the understanding of the human language capacity, as studied by linguistics and cognitive science.  We outline the evolution of NLP from its beginnings until the age of large language models, and highlight for each of its main paradigms some similarities and differences with theories of the human language capacity.  We conclude that the evolution of language technology has not substantially deepened our understanding of how human minds process natural language, despite the impressive language abilities attained by current chatbots using artificial neural networks. 
\end{abstract}

\section{Introduction}
\label{sec:introduction}

If you know how to build a clock, this means that you understand how to measure time. If you know how to build a mechanical calculator, it means that you understand how to perform arithmetical calculations. But if you know how to create a chatbot who can answer your questions almost like a human, do you necessarily understand how human language works?

Computer programs that can use human language have been imagined as early as the first computers, aiming for instance to speed up document search \citep{bush1945,bush1996reprint} or to translate texts \citep{weaver1949,weaver1955reprint}. But what kind of knowledge is needed to accomplish these tasks?  We may think that a formal or algorithmic description of the language to be processed is required, or, even better, a general theory of human language(s). Implementing such a description into a computer program would then constitute a language processing system with capacities similar to those of human speakers.

However, the history of the field traditionally called `natural language processing' has followed quite a different evolution. Early on, the path taken by theoretical language analysis diverged significantly from the path taken by computer applications. ``Whenever I fire a linguist our system performance improves,'' a team leader reportedly stated in the 1980s \citep{jelinek2005}. The science and the technology of human language have steadily grown apart, to the point that today, what we call `large language models' (LLM) bear only a distant relationship to cognitive models of human language.  This paper aims to retrace how we got there, and to discuss some differences but also some similarities between LLMs and the processing of human language in the brain.

\section{From Rules to Statistical Models}

Starting in the 1950s, numerous connections have been made between the  disciplines of formal linguistics and natural language processing. The foundations of a hierarchical model of how humans represent and process language were laid during the 1960s and 1970s. Two types of levels are traditionally considered for analysis \citep{jackendoff2002}. First, one can distinguish, in written language, the levels of sounds, morphemes (parts of words such as prefixes), words, phrases, simple sentences, complex sentences, and texts. Then, for several of these levels, one can study separately or jointly their structural vs.\ semantic vs.\ pragmatic properties.\footnote{While semantics studies the standalone meaning of words or sentences \citep{chierchia2000meaning}, pragmatics considers meaning in context \citep{levinson1983pragmatics,sperber1986relevance}.}  For instance, at the word level, structural properties include word formation and inflection rules \citep{haspelmath2013understanding}; at the sentence level, syntax \citep{chomsky1957syntactic,carnie2021syntax}; and at the text level, discourse structure and relations \citep{mann1988rst,sanders2021unifying}.  Although the representation of these levels in human brains remains largely to be discovered, many specialized areas have been identified. And several studies have found evidence that some of these levels are differentiated also in recent artificial neural networks that process human language \citep{tenney-etal-2019-bert}.

Hierarchical models of human language processing were the first blueprint for the design of automated processes that could reproduce these steps with computer programs.
For example, to build an automated translation system (a concept known as `machine translation' since the 1950s), the idea arose of combining a morphological analyzer with a syntactic analyzer, and then a semantic one, in order to construct a logical representation of the meaning of a source sentence. To generate the translation, this logical form was used to build a syntactic tree in the target language, with the appropriate words and their inflections.  This model is the one with the most abstract repreesentation in Vauquois' triangle. The challenges posed by such an approach were numerous, both in terms of representation formalisms and in terms of implementation.  How, indeed, could all the necessary knowledge be provided? The cost of developing such systems was very high, and the simplified models that were created did not help in understanding the human capacity for translation \citep{hutchins1986machine,nirenburg2003readings}.

Another area where the challenge of explicitly formalizing knowledge seemed impossible to solve was the transcription of spoken language into written text (a task known as `automatic speech recognition' or ASR). The variability of speakers and the mutual influence between consecutive sounds in speech were serious challenges in the 1980s. Gradually, the idea of a statistical model gained traction, combining the following modules to provide a trainable ASR system. A first module converts the audio signal into the probability that certain phonemes have been uttered.  Then, a second module estimates the probability that a certain sequence of sounds, corresponding to a certain sequence of letters, has been uttered.  A third module, called a `language model', estimates the  probability of various word-sequence candidates and selects the most likely one as the reesult of ASR \citep{rabiner2002tutorial}.

From the 1990s onward, statistical language models (in the sense we just defined) learned automatically from text data dominated those based on linguistic rules \citep[for translation, see e.g.][]{Koehn2009}. Their principle consisted of memorizing the frequencies of word sequences (e.g., between 1 and 5 words) seen in a large set of texts and using them to estimate the probability of a new word sequence by decomposing it into sequences that has already been memorized during training. It is difficult to see in this process and in the model itself any analogy with human linguistic abilities, which go far beyond judgments about the probabilities of sentences.

\begin{figure*}[ht]
\centering
\subfloat[An autoencoder network can be trained to reproduce the activations from the input layer in its output layer.  This relies on the emergence, in the hidden layer, of a low-dimensional representation of an input vector, materialized by the activations of units in the hidden layer.]{\label{fig:1a}
\centering
\includegraphics[width=0.47\linewidth]{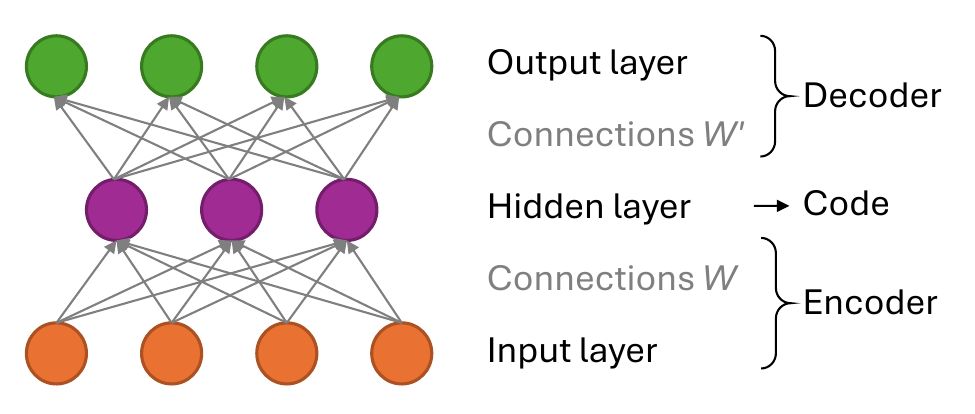}
}\hfill
\subfloat[A recurrent network has successive states, corresponding for instance to the processing of a sequence of words given as input. The activations of the units in the hidden layer depend on their activations of the input layer, but also on those of the previous state of the hidden layer. The network can be trained so that the output vector generates predictions of words that reasonably continue the input sequence.]{\label{fig:1b}
\centering
\includegraphics[width=0.47\linewidth]{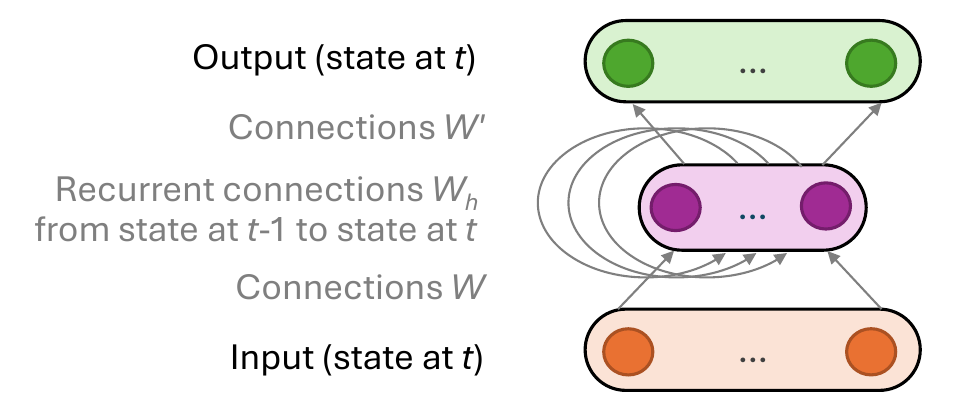}
}
\caption{Two simple structures of artificial neural networks that have been used to designed early neural language models. The core idea of such networks is that the activation of each unit is a function (often non linear) of the weighted sum of the activations of units connected to it.}
\label{fig:1}
\end{figure*}

\section{Language Processing with Neural Networks}

In the early 2000s, advances in the design of artificial neural networks for image analysis encouraged their application to human language as well. This led to the idea of representing the meaning of words, and then of sentences, using low-dimensional dense vectors called `embeddings' (typically, several hundred non-zero values).  What are good vectors? They are such that words or sentences with similar meanings are represented by vectors with similar directions, as measured by the cosine of the angle between them. 

To automatically construct word embeddings, one of the first and simplest models is word2vec \citep{NIPS2013_9aa42b31}, an architecture inspired by an autoencoder network, as sketched in Figure~\ref{fig:1a}. Entering a word into the word2vec network means activating a specific input unit, which then triggers the activation of one or more hidden layers, up to a low-dimensional central layer. The activations in this layer are the low-dimensional dense vector which is the embedding of the input word. In plain autoencoders, training consists of using this embedding to reproduce the input as accurately as possible in the output layer.  However, the successful training of the word2vec model is rather based on predicting, in the output layer, several likely neighboring words of the input word, as this task is easily formulated with training data derived from large quantities of text. 

Many other types of networks emerged during the 2000s and 2010s, such as the recurrent neural networks represented schematically in Figure~\ref{fig:1b}, originally introduced by \citet{elman1990}. Training these networks was also based on the language modeling task, i.e.\ aiming to a assign a probability to a sequence of words, or to predict the words that can reasonably complete a given sequence.  When the sequence is made of a source and a target sentence, separated by a `translate' symbol, this approach was applied to the design of neural networks for machine translation with a certain success \citep{cho-etal-2014-learning}, especially when enhanced with an attention model \citep{bahdanau2014neural,luong-etal-2015-effective}.

One might legitimately think that learning with neural networks looks like magic: how is it possible to start from the word prediction task and obtain a network that has learned word embeddings which obey the principle that ``similar words (or sentences) have similar vectors?''  A sketchy explanation is that, in order to correctly complete a sequence, it is important to identify classes of similarity for words, and appropriate structures of neural networks constrain them to group embeddings in a low-dimensional space according to such classes. Since 2017, for language processing, one structure of neural networks has essentially replaced all others, namely the Transformer system.

\section{The Transformer}

The invention of the Transformer architecture by a team from Google in 2017 was arguably the most significant advance in neural networks of the past ten years \citep{NIPS2017_3f5ee243}. Dissatisfied with the difficulties of training recurrent neural networks for machine translation, the authors searched for a new architecture that allows for parallel processing.  Inspired from image processing, and instead of computing the embedding of each word based on previous ones, a Transformer neural network starts with a non-contextual embedding for each word of a sentence and transforms each one based on all the others.\footnote{Technically, the Transformer networks do not operate with words, but with tokens, which are drawn from a fixed vocabulary of words, subwords, and letters \citep{sennrich-etal-2016-neural,kudo-richardson-2018-sentencepiece}.}  The learned transformation steers vectors away from their \textit{a priori} value for each word, and directs them in coherent directions based on their context, i.e.\ on the other vectors of the text.  Moreover, by stacking dozens of Transformer blocks, the contextualized embeddings of input words gradually evolve towards a representation of meaning that abstracts away from surface form. In the original Transformer system, it is the `encoder' part of the network computes these embeddings for the sentence to be translated. The `decoder' part uses almost the same architecture, except that only the previous words can be used in the transformation process. Its role is to generate, word by word, the translated sentence: in other words, it is an auto-regressive language model conditioned on the encoder's representations. The translation is then obtained as the most probable word sequence, with various techniques to approach the optimal sequence.

While Google focused on developing encoders \citep{devlin-etal-2019-bert}, a team at OpenAI turned its attention to decoders from 2018 on, and demonstrated that the controlled increase of their sizes leads to new or increased capabilities for text generation, question answering, and problem solving. The team solved challenges related to training very large neural networks, leading to the models of the GPT family (Generative Pretrained Transformer). In 2019, GPT-2 distinguished itself by its ability to generate high-quality fictional articles from a title and a prompt \citep{radford2019language}. However, the large amounts of training data used by GPT-2 and then by GPT-3 \citep{NEURIPS2020_1457c0d6} were not sufficient to teach them that a question must be completed by an answer, and more generally that instructions must be completed by their execution -- and not, for instance, by their reformulation. To address this issue, OpenAI researchers applied their previous work on general reinforcement learning to the problem of language model adaptation \citep{10.5555/3045118.3045319,schulman2017proximal}.  Around 2022, they created an automated preference model which predicts human ratings of GPT responses, and used this model to perform reinforcement learning on a GPT decoder \citep{NEURIPS2022_b1efde53}. With the help of tens of thousands of instruction/answer pairs created by humans, GPT-3.5 (also referred to as InstructGPT) was able to learn to respond cooperatively, while also following ethical principles in its answers, thus denying to answer questions about unethical matters such as synthesizing explosives or drugs, or planning illegal actions, despite its initial ability to answer them. 

This method enabled the public launch of ChatGPT at the end of 2022, with the success we know today.  Several other large language models (LLMs) followed suit, e.g.\ Anthropic's Claude series, Google's Gemma and Gemini, Meta's Llama series, Microsoft's Phi, as well as DeepSeek-R1, Mistral, Qwen, and many others.\footnote{A \href{https://huggingface.co/models?pipeline_tag=text-generation&sort=likes}{repository of open-weight models} is provided by Hugging Face.}

\section{Generative AI and the Human Language Capacity}

Beyond their popularity and challenges to many aspects of contemporary societies, what do Generative AI models tell us about the human language capacity?  First, we can observe that humans often judge these models based on their use of natural language for solving problem. LLMs possess an incredible amount of knowledge and use it effectively to answer questions, not just by simply quoting from their training data, but by combining relevant pieces of knowledge and generalizing from known examples. Even though LLMs are tireless when it comes to solving tasks within their reach, they are often accused of relying primarily on their memory, without demonstrating genuine creativity.

Unlike human beings, the knowledge that LLMs possess appears to be disembodied: it is not based on sensorimotor capabilities. Consequently, LLMs probably do not have a model of the physical world similar to that of living beings, although little is known yet about their internal representations. When it comes to connecting LLMs to sensors and actuators, for instance in order to build a humanoid robot or an autonomous car, many fundamental problems remain to be solved.

Moreover, the process of language learning in LLMs is radically different from how human children learn language.  While LLMs learn essentially through text completion tasks, followed by reinforcement learning guided by an artificial reward, children centrally benefit from constant interactions with other speakers. Furthermore, while children in their first years hear at most a few thousand words per day, the volume of training data for the largest LLMs is the size of the greatest public libraries.  Similarly, the energy that is necessary to train an LLM is measured in GWh, whereas a human brain consumes less than one kWh per day, which is a million times smaller, even when taking into account the difference in training time (months vs.\ years).

These essential differences in the structure and training of LLMs vs.\ humans indicate that they are for now a poor model of the human language function.  Aside from the concept of formal neurons and the propagation of activations via connections, no element of the Transformer exhibits a strong analogy with the human brain. The specialization of brain areas for language, the variety of types of synapses, the action potentials triggering neural spikes -- none of these elements can be found in the Transformer.

\section{Conclusion}

A stunning series of revolutions has swept through the field of natural language processing in the past twenty years. Machine translation itself, with many popular online systems, has witnessed three paradigm shifts: from statistical models, to RNNs with attention, to Transformer-based encoder-decoders, and now to LLMs.

Who could have predicted, even at the beginning of 2022, the conversational level reached by ChatGPT at the end of the same year, and soon by the other competing models?  Although their structure and learning mechanisms are well understood, the relevance of their answers is intriguing. Do they truly understand human language? It all depends on the criteria defining understanding, but conversations with an AI chatbot can certainly be more interesting than those we have with some human fellows. And while errors and hallucinations still appear in LLM generated texts, aren't they also frequent in human productions? As for the connectionist approach, despite its fundamental differences with human brains, doesn't it share with them a similar mode of memorization, based on the weights of connections, and a certain robustness when these weights are modified?

Humans beings may not be the only ones understanding language: LLMs may do so as well, in their own ways.  Does this help us understand how human language works? Ironically, the better these models emulate the human capacity for using language, the less their creators seem to understand how that is possible, and how structural changes in these models could help them to perform even better. It seems that We have created speaking artifacts through a series of evolutions of computational systems, after many trials and errors: a curious and maybe unintended imitation of the natural evolution of cognition.

\section*{Acknowledgments}

This is a translated and revised version of the article ``\href{https://www.lajauneetlarouge.com/agents-conversationnels-et-comprehension-du-langage-humain/}{Agents conversationnels et compréhension du langage humain}'' published in \textit{La Jaune et la Rouge, magazine des alumni de l'École Polytechnique}, n.\ 810, p.\ 59-62, December 2025.  I would like to thank Franck Ramus for inviting me to write this text for the special issue he coordinated and Sandrine Zufferey for her suggestions on a first version of the text.

\bibliography{references}

@article{bahdanau2014neural,
  title={Neural Machine Translation by Jointly Learning to Align and Translate},
  author={Bahdanau, Dzmitry and Cho, KyungHyun and Bengio, Yoshua},
  journal={arXiv preprint arXiv:1409.0473},
  year={2014},
  url={https://arxiv.org/abs/1409.0473}
}

@inproceedings{NEURIPS2020_1457c0d6,
 author = {Brown, Tom and Mann, Benjamin and Ryder, Nick and Subbiah, Melanie and Kaplan, Jared D and Dhariwal, Prafulla and Neelakantan, Arvind and Shyam, Pranav and Sastry, Girish and Askell, Amanda and Agarwal, Sandhini and Herbert-Voss, Ariel and Krueger, Gretchen and Henighan, Tom and Child, Rewon and Ramesh, Aditya and Ziegler, Daniel and Wu, Jeffrey and Winter, Clemens and Hesse, Chris and Chen, Mark and Sigler, Eric and Litwin, Mateusz and Gray, Scott and Chess, Benjamin and Clark, Jack and Berner, Christopher and McCandlish, Sam and Radford, Alec and Sutskever, Ilya and Amodei, Dario},
 booktitle = {Advances in Neural Information Processing Systems},
 editor = {H. Larochelle and M. Ranzato and R. Hadsell and M.F. Balcan and H. Lin},
 pages = {1877--1901},
 publisher = {Curran Associates, Inc.},
 title = {Language Models are Few-Shot Learners},
 url = {https://proceedings.neurips.cc/paper_files/paper/2020/file/1457c0d6bfcb4967418bfb8ac142f64a-Paper.pdf},
 volume = {33},
 year = {2020}
}

@article{bush1945,
    author = {Bush, Vannevar},
    title = {As We May Think},
    journal = {The Atlantic Monthly},
    month = {July},
    year = {1945},
    pages = {101-108}
}

@article{bush1996reprint,
    author = {Bush, Vannevar},
    title = {As We May Think (reprint)},
    journal = {ACM Interactions},
    volume = {3},
    number = {2},
    month = {March},
    year = {1996},
    url = {https://dl.acm.org/doi/10.1145/227181.227186},
    pages = {35-46},
    publisher = {Association for Computing Machinery}
}

@book{carnie2021syntax,
  title={Syntax: A generative introduction},
  author={Carnie, Andrew},
  year={2021},
  publisher={John Wiley \& Sons}
}

@book{chierchia2000meaning,
  title={Meaning and grammar: An introduction to semantics},
  author={Chierchia, Gennaro and McConnell-Ginet, Sally},
  year={2000},
  publisher={MIT press Cambridge, MA}
}

@inproceedings{cho-etal-2014-learning,
    title = "Learning Phrase Representations using {RNN} Encoder{--}Decoder for Statistical Machine Translation",
    author = {Cho, Kyunghyun  and
      van Merri{\"e}nboer, Bart  and
      Gulcehre, Caglar  and
      Bahdanau, Dzmitry  and
      Bougares, Fethi  and
      Schwenk, Holger  and
      Bengio, Yoshua},
    editor = "Moschitti, Alessandro  and
      Pang, Bo  and
      Daelemans, Walter",
    booktitle = "Proceedings of the 2014 Conference on Empirical Methods in Natural Language Processing ({EMNLP})",
    month = oct,
    year = "2014",
    address = "Doha, Qatar",
    publisher = "Association for Computational Linguistics",
    url = "https://aclanthology.org/D14-1179/",
    doi = "10.3115/v1/D14-1179",
    pages = "1724--1734"
}

@book{chomsky1957syntactic,
  title={Syntactic structures},
  author={Chomsky, Noam},
  year={2000},
  publisher={Mouton},
  address = {The Hague}
}

@inproceedings{devlin-etal-2019-bert,
    title = "{BERT}: Pre-training of Deep Bidirectional Transformers for Language Understanding",
    author = "Devlin, Jacob  and
      Chang, Ming-Wei  and
      Lee, Kenton  and
      Toutanova, Kristina",
    editor = "Burstein, Jill  and
      Doran, Christy  and
      Solorio, Thamar",
    booktitle = "Proceedings of the 2019 Conference of the North {A}merican Chapter of the Association for Computational Linguistics: Human Language Technologies, Volume 1 (Long and Short Papers)",
    month = jun,
    year = "2019",
    address = "Minneapolis, Minnesota",
    publisher = "Association for Computational Linguistics",
    url = "https://aclanthology.org/N19-1423/",
    doi = "10.18653/v1/N19-1423",
    pages = "4171--4186"
}

@article{elman1990,
author = {Elman, Jeffrey L.},
title = {Finding Structure in Time},
journal = {Cognitive Science},
volume = {14},
number = {2},
pages = {179-211},
doi = {https://doi.org/10.1207/s15516709cog1402\_1},
url = {https://onlinelibrary.wiley.com/doi/abs/10.1207/s15516709cog1402_1},
year = {1990}
}

@book{haspelmath2013understanding,
  title={Understanding Morphology},
  author={Haspelmath, Martin and Sims, Andrea},
  year={2013},
  publisher={Routledge},
  address = {London, UK}
}

@book{hutchins1986machine,
  title={Machine translation: past, present, future},
  author={Hutchins, William John},
  year={1986},
  publisher={Ellis Horwood},
  address = {Chichester, UK}
}

@book{jackendoff2002,
    author = {Ray Jackendoff},
    title = {Foundations of language: brain, meaning, grammar, evolution},
    publisher = {Oxford University Press},
    year = {2002}
}

@article{jelinek2005,
    author = {Jelinek, Frederick},
    title = {Some of my Best Friends are Linguists},
    journal = {Language Resources and Evaluation},
    volume = {39},
    number = {1},
    pages = {25-34},
    year = {2005},
    url = {https://doi.org/10.1007/s10579-005-2693-4}
}

@book{Koehn2009, 
address={Cambridge, UK}, 
title={Statistical Machine Translation}, 
publisher={Cambridge University Press}, 
author={Koehn, Philipp}, 
year={2009}
}

@inproceedings{kudo-richardson-2018-sentencepiece,
    title = "{S}entence{P}iece: A simple and language independent subword tokenizer and detokenizer for Neural Text Processing",
    author = "Kudo, Taku  and
      Richardson, John",
    editor = "Blanco, Eduardo  and
      Lu, Wei",
    booktitle = "Proceedings of the 2018 Conference on Empirical Methods in Natural Language Processing: System Demonstrations",
    month = nov,
    year = "2018",
    address = "Brussels, Belgium",
    publisher = "Association for Computational Linguistics",
    url = "https://aclanthology.org/D18-2012/",
    doi = "10.18653/v1/D18-2012",
    pages = "66--71"
}

@book{levinson1983pragmatics,
  title={Pragmatics},
  author={Levinson, Stephen C},
  year={1983},
  publisher={Cambridge University Press},
  address={Cambridge, UK}
}

@inproceedings{luong-etal-2015-effective,
    title = "Effective Approaches to Attention-based Neural Machine Translation",
    author = "Luong, Thang  and
      Pham, Hieu  and
      Manning, Christopher D.",
    editor = "M{\`a}rquez, Llu{\'i}s  and
      Callison-Burch, Chris  and
      Su, Jian",
    booktitle = "Proceedings of the 2015 Conference on Empirical Methods in Natural Language Processing",
    month = sep,
    year = "2015",
    address = "Lisbon, Portugal",
    publisher = "Association for Computational Linguistics",
    url = "https://aclanthology.org/D15-1166/",
    doi = "10.18653/v1/D15-1166",
    pages = "1412--1421"
}

@article{mann1988rst,
url = {https://doi.org/10.1515/text.1.1988.8.3.243},
title = {Rhetorical Structure Theory: Toward a functional theory of text organization},
author = {Mann, William C. and Thompson, Sandra A.},
pages = {243--281},
volume = {8},
number = {3},
journal = {Text},
year = {1988}
}

@inproceedings{NIPS2013_9aa42b31,
 author = {Mikolov, Tomas and Sutskever, Ilya and Chen, Kai and Corrado, Greg S and Dean, Jeff},
 booktitle = {Advances in Neural Information Processing Systems},
 editor = {C.J. Burges and L. Bottou and M. Welling and Z. Ghahramani and K.Q. Weinberger},
 pages = {},
 publisher = {Curran Associates, Inc.},
 title = {Distributed Representations of Words and Phrases and their Compositionality},
 url = {https://proceedings.neurips.cc/paper_files/paper/2013/file/9aa42b31882ec039965f3c4923ce901b-Paper.pdf},
 volume = {26},
 year = {2013}
}

@book{nirenburg2003readings,
  title={Readings in Machine Translation},
  author={Nirenburg, Sergei and Somers, Harold L. and Wilks, Yorick},
  year={2003},
  publisher={MIT Press},
  address = {Cambridge, MA, USA}
}

@inproceedings{NEURIPS2022_b1efde53,
 author = {Ouyang, Long and Wu, Jeffrey and Jiang, Xu and Almeida, Diogo and Wainwright, Carroll and Mishkin, Pamela and Zhang, Chong and Agarwal, Sandhini and Slama, Katarina and Ray, Alex and Schulman, John and Hilton, Jacob and Kelton, Fraser and Miller, Luke and Simens, Maddie and Askell, Amanda and Welinder, Peter and Christiano, Paul F and Leike, Jan and Lowe, Ryan},
 booktitle = {Advances in Neural Information Processing Systems},
 editor = {S. Koyejo and S. Mohamed and A. Agarwal and D. Belgrave and K. Cho and A. Oh},
 pages = {27730--27744},
 publisher = {Curran Associates, Inc.},
 title = {Training language models to follow instructions with human feedback},
 url = {https://proceedings.neurips.cc/paper_files/paper/2022/file/b1efde53be364a73914f58805a001731-Paper-Conference.pdf},
 volume = {35},
 year = {2022}
}

@article{rabiner2002tutorial,
  title={A tutorial on hidden {M}arkov models and selected applications in speech recognition},
  author={Rabiner, Lawrence R.},
  journal={Proceedings of the IEEE},
  volume={77},
  number={2},
  pages={257--286},
  year={2002},
  publisher={Ieee}
}

@article{radford2019language,
  title={Language models are unsupervised multitask learners},
  author={Radford, Alec and Wu, Jeffrey and Child, Rewon and Luan, David and Amodei, Dario and Sutskever, Ilya and others},
  journal={OpenAI blog},
  volume={1},
  number={8},
  pages={9},
  year={2019},
  url = {https://cdn.openai.com/better-language-models/language_models_are_unsupervised_multitask_learners.pdf}
}

@article{sanders2021unifying,
url = {https://doi.org/10.1515/cllt-2016-0078},
title = {Unifying dimensions in coherence relations: How various annotation frameworks are related},
author = {Ted J. M. Sanders and Vera Demberg and Jet Hoek and Merel C. J. Scholman and Fatemeh Torabi Asr and Sandrine Zufferey and Jacqueline Evers-Vermeul},
pages = {1--71},
volume = {17},
number = {1},
journal = {Corpus Linguistics and Linguistic Theory},
doi = {doi:10.1515/cllt-2016-0078},
year = {2021}
}

@inproceedings{10.5555/3045118.3045319,
    author = {Schulman, John and Levine, Sergey and Moritz, Philipp and Jordan, Michael and Abbeel, Pieter},
    title = {Trust region policy optimization},
    year = {2015},
    publisher = {JMLR.org},
    booktitle = {Proceedings of the 32nd International Conference on Machine Learning - Volume 37},
    pages = {1889–1897},
    numpages = {9},
    location = {Lille, France},
    series = {ICML'15}
    }

@article{schulman2017proximal,
  title={Proximal policy optimization algorithms},
  author={Schulman, John and Wolski, Filip and Dhariwal, Prafulla and Radford, Alec and Klimov, Oleg},
  journal={arXiv preprint arXiv:1707.06347},
  url={https://arxiv.org/abs/1707.06347},
  year={2017}
}

@inproceedings{sennrich-etal-2016-neural,
    title = "Neural Machine Translation of Rare Words with Subword Units",
    author = "Sennrich, Rico  and
      Haddow, Barry  and
      Birch, Alexandra",
    editor = "Erk, Katrin  and
      Smith, Noah A.",
    booktitle = "Proceedings of the 54th Annual Meeting of the Association for Computational Linguistics (Volume 1: Long Papers)",
    month = aug,
    year = "2016",
    address = "Berlin, Germany",
    publisher = "Association for Computational Linguistics",
    url = "https://aclanthology.org/P16-1162/",
    doi = "10.18653/v1/P16-1162",
    pages = "1715--1725"
}

@book{sperber1986relevance,
  title={Relevance: Communication and Cognition},
  author={Sperber, Dan and Wilson, Deirdre},
  volume={142},
  year={1986},
  publisher={Harvard University Press Cambridge, MA}
}

@inproceedings{tenney-etal-2019-bert,
    title = "{BERT} Rediscovers the Classical {NLP} Pipeline",
    author = "Tenney, Ian  and
      Das, Dipanjan  and
      Pavlick, Ellie",
    editor = "Korhonen, Anna  and
      Traum, David  and
      M{\`a}rquez, Llu{\'i}s",
    booktitle = "Proceedings of the 57th Annual Meeting of the Association for Computational Linguistics",
    month = jul,
    year = "2019",
    address = "Florence, Italy",
    publisher = "Association for Computational Linguistics",
    url = "https://aclanthology.org/P19-1452/",
    doi = "10.18653/v1/P19-1452",
    pages = "4593--4601"
}

@inproceedings{NIPS2017_3f5ee243,
 author = {Vaswani, Ashish and Shazeer, Noam and Parmar, Niki and Uszkoreit, Jakob and Jones, Llion and Gomez, Aidan N and Kaiser, {\L}ukasz and Polosukhin, Illia},
 booktitle = {Advances in Neural Information Processing Systems},
 editor = {I. Guyon and U. Von Luxburg and S. Bengio and H. Wallach and R. Fergus and S. Vishwanathan and R. Garnett},
 pages = {},
 publisher = {Curran Associates, Inc.},
 title = {Attention is All you Need},
 url = {https://proceedings.neurips.cc/paper/2017/file/3f5ee243547dee91fbd053c1c4a845aa-Paper.pdf},
 volume = {30},
 year = {2017}
}

@article{weaver1949,
    author = {Warren Weaver},
    title = {Translation},
    journal = {Unpublished Memorandum},
    month = {July},
    year = {1949}
}

@incollection{weaver1955reprint,
    author = {Warren Weaver},
    title = {Translation},
    booktitle = {Machine Translation of Languages},
    publisher = {The MIT Press},
    address = {Cambridge, MA, USA},
    year = {1955},
    pages = {15-23},
    editor = {Locke, W.N. and Booth, D.A.}
}

\end{document}